\documentclass[runningheads]{llncs}

\usepackage{iciap}
\usepackage{iciapabbrv}

\usepackage{graphicx}
\usepackage{booktabs}

\usepackage[accsupp]{axessibility}  % Improves PDF readability for those with disabilities.

\usepackage{hyperref}

\usepackage{orcidlink}
\usepackage{verbatim}
\usepackage{pifont}
\usepackage{multirow} 
\usepackage{xspace}
\usepackage{booktabs}
\usepackage{xcolor}

\usepackage{amsmath}
\usepackage{multicol}
\usepackage{amssymb}
\usepackage{makecell}

\newcommand{\method}{MAD\textsc{POT}\xspace}
\newcommand{\myvspace}{\vspace{0.5em}}

\begin{document}

\title{MADPOT: Medical Anomaly Detection with CLIP Adaptation and Partial Optimal Transport} 

\titlerunning{MADPOT}

%%\author{First Author\inst{1}\orcidlink{0000-1111-2222-3333} \and
%%Second Author\inst{2,3}\orcidlink{1111-2222-3333-4444} \and
%%Third Author\inst{3}\orcidlink{2222--3333-4444-5555}}

%%\authorrunning{F.~Author et al.}

%%\institute{Princeton University, Princeton NJ 08544, USA \and
%%Springer Heidelberg, Tiergartenstr.~17, 69121 Heidelberg, Germany
%%\email{lncs@springer.com}\\
%%\url{http://www.springer.com/gp/computer-science/lncs} \and
%%ABC Institute, Rupert-Karls-University Heidelberg, Heidelberg, Germany\\
%%\email{\{abc,lncs\}@uni-heidelberg.de}}

\author{Mahshid Shiri\inst{1}  \and
Cigdem Beyan\inst{1} \thanks{Corresponding author} \and
Vittorio Murino\inst{1,2}}
% index{Shiri, Mahshid}
% index{Beyan, Cigdem}
% index{Murino, Vittorio
\authorrunning{M. Shiri et al.}

\institute{
Department of Computer Science, University of Verona, Verona, Italy \and
AI for Good (AIGO) Research Unit, Istituto Italiano di Tecnologia, Genoa, Italy \\
\email{\{mahshid.shiri, cigdem.beyan, vittorio.murino\}@univr.it}
}

\maketitle

\begin{abstract}
 Medical anomaly detection (AD) is challenging due to diverse imaging modalities, anatomical variations, and limited labeled data. We propose a novel approach combining visual adapters and prompt learning with Partial Optimal Transport (POT) and contrastive learning (CL) to improve CLIP’s adaptability to medical images, particularly for AD. Unlike standard prompt learning, which often yields a single representation, our method employs multiple prompts aligned with local features via POT to capture subtle abnormalities. CL further enforces intra-class cohesion and inter-class separation. Our method achieves state-of-the-art results in few-shot, zero-shot, and cross-dataset scenarios without synthetic data or memory banks. The code is available at \url{https://github.com/mahshid1998/MADPOT}.

%\keywords{First keyword  \and Second keyword \and Another keyword.}
\end{abstract}

\section{Introduction}
Medical anomaly detection (AD) aims to identify unusual patterns in medical data, a task made challenging by the absence of a universal definition of anomaly and the considerable variability in medical images, both across different modalities and anatomical regions \cite{huang2024adapting,chen2023zero,zhang2020viral}. Ideally, a model should exhibit versatility across diverse data types, regardless of variations in anomaly size, modality, or underlying anatomical structures. Additionally, it should be capable of handling tasks involving few-shot learning (where only a limited number of labeled examples are available for training, e.g., \cite{ding2022catching,huang2022registration}) or zero-shot learning (where the model is trained on one set of modalities but tested on completely unseen modalities, e.g., \cite{huang2024adapting}).

Recent advancements in large-scale pre-trained visual-language models (VLMs) have led to more robust and generalizable AD. Notably, CLIP \cite{radford2021learning} has been adapted by several studies, such as \cite{chen2023zero,cao2024adaclip}, for industrial defect detection. Similarly, in the field of medical AD, which involves anomaly classification (AC, i.e., binary classification of normal vs. abnormal) and anomaly segmentation (AS), CLIP-based architectures have also been effectively applied \cite{zhang2024mediclip,huang2024adapting}.
These studies \cite{zhang2024mediclip,huang2024adapting} have demonstrated that relying solely on CLIP \cite{radford2021learning}, which is pre-trained on natural images, is insufficient for medical AD. This limitation arises not only from the domain gap between natural and medical images, but also from CLIP's original training, which primarily focuses on aligning the class semantics of foreground objects. As a result, its ability to generalize and capture subtle visual abnormalities in medical AD is constrained. 

Consequently, CLIP can be leveraged through fine-tuning on medical images or by adapting the visual encoder, which primarily captures image semantics, to better align with the unique characteristics of medical data. For instance, MVFA \cite{huang2024adapting} utilizes adapters, designed as linear layers, which are attached to different layers of the CLIP vision encoder, with training performed for both AC and AS tasks. That approach \cite{huang2024adapting} is effective for both few-shot and zero-shot settings. In contrast, MediCLIP \cite{zhang2024mediclip} uses convolutional layer adapters but focuses solely on few-shot AD, requiring a significant amount of synthesized medical image anomaly instances to achieve optimal performance. Another approach to adapting CLIP is through text-based adaptation via prompt engineering, where fixed hand-crafted prompts (e.g., ``a black and white photo of a normal [brain]''), as applied in \cite{huang2024adapting}, can be used. Alternatively, prompt learning can be employed to align the textual representations more closely with task-specific language, i.e., medical terminology and anomaly-related descriptors. This approach is implemented through CoOP \cite{zhou2022coop} in the few-shot medical AD pipeline of \cite{zhang2024mediclip}.

In this study, we combine both visual adapters and prompt learning. We argue that, since anomaly descriptions may share similarities across different datasets, incorporating textual information helps reduce reliance solely on visual data. This approach could enhance model performance, especially in data-scarce scenarios such as few-shot or zero-shot learning, where the test time modality is completely unseen.
On the other hand, while learned prompts can potentially improve upon manually crafted ones, they may still fall short of fully representing a class that possesses multiple attributes and complex contextual relationships. For example, we can describe the class ``Lung Tumor'' from various perspectives, such as the size and shape of the tumor, its location within the lung, the density of surrounding tissue, and the presence of any vascular invasion or nearby organ involvement. This motivates us to learn multiple prompts. However, this is akin to matching the mean of the prompt features with the visual features, as all prompts are pulled toward a single point, resulting in them learning the same characteristic \cite{chen2022plot}. This undermines the goal of learning diverse and comprehensive prompts. 

To overcome this limitation, we target prompt learning with Optimal Transport (OT) \cite{villani2008optimal}, where OT is used to align local visual features with multiple textual prompts through a learned transport plan \cite{chen2022plot}.
However, we argue that the classical OT problem may present two key limitations for medical AD. First, it imposes a balance condition, requiring the total ``mass'' (i.e., the summed probability) of the source and target distributions to be equal \cite{koehl2021physics}. This constraint can be limiting when analyzing medical images, especially for detecting small, localized anomalies that constitute only a minor portion of the data. Second, the OT plan does not explicitly preserve local discriminative structures, such as groups or clusters of features, which we consider important for distinguishing between classes in classification tasks \cite{wang2024probability}.

\begin{figure}[t!]
    \centering
    \includegraphics[width=1\linewidth]{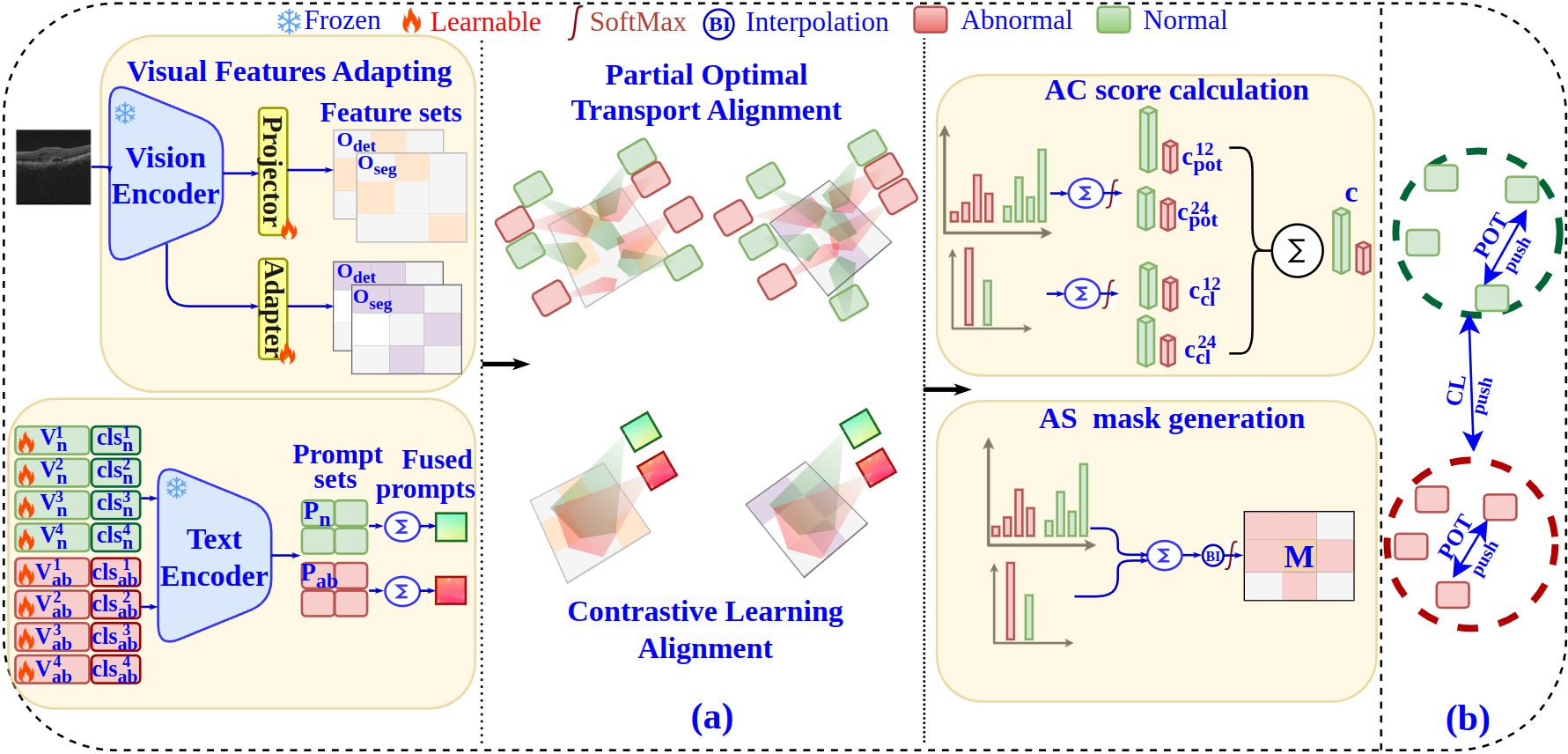}
    \caption{(a) Overview of \method:  
    Visual features are extracted via an adapter and projection on the CLIP vision encoder. Separate prompt sets for normal and abnormal are learned and fused into class-specific prompts.  POT produces multiple logits, and CL is applied to the visual features and fused prompts. The final score is the average of the logits from POT and CL.
    (b) POT enables the model to capture localized features, while CL encourages the prompts to differentiate between normal and abnormal classes. By combining CL and POT, we maximize multimodal similarity within each class and minimize it across classes, preserving the diversity of learned prompts within each class.}
    \label{im:proposed}
\end{figure}

To address the first limitation, we employ Partial Optimal Transport (POT), which transports only a fixed fraction of the total mass. This allows us to guide the alignment toward the most informative image patches, typically small and subtle anomalies, while downplaying irrelevant or healthy regions. This selective focus is crucial in medical AD, where abnormalities often occupy only a minor portion of the image. To address the second limitation, we integrate contrastive learning (CL) with POT, enabling the model to explicitly learn local discriminative structures. While POT guides alignment toward informative regions, CL promotes the formation of well-separated clusters by pulling similar features closer and pushing dissimilar ones apart. This helps preserve meaningful intra-class and inter-class distinctions, which are essential for classification tasks in medical AD (see Fig. \ref{im:proposed}b). 

Our method, called \method, demonstrates superior performance in medical AD across few-shot, zero-shot, and cross-dataset settings (i.e., training and testing occur on the same modality but different datasets), surpassing all state-of-the-art (SOTA) methods. Additionally, our pipeline eliminates the need for memory banks, as seen in e.g., \cite{huang2024adapting}, or additional synthetic data, as used in \cite{zhang2024mediclip}. \\

\noindent {The main contributions of our work are:}

\noindent \textbf{(1)} The pioneering investigation of prompt learning with OT and POT and their novel integration with CL in medical AD, particularly in few-shot and zero-shot settings. This is the first exploration of these techniques in this context, highlighting their potential for reducing data dependency and enhancing model flexibility.

\noindent \textbf{(2)} A comprehensive evaluation of OT, POT, CL, and vision adapters, alongside their various combinations, providing an in-depth analysis of how these components contribute to model performance across datasets with substantial modality and distribution differences.

\noindent \textbf{(3)} Demonstration of strong generalization and noticeable performance improvements, validated through cross-dataset and cross-modality evaluations. Our method shows robustness across diverse medical modalities and anatomical regions, making it a possible, reliable solution for real-world clinical applications.

\section{Related Works}
        \noindent \textbf{Anomaly Detection.}
                The generic AD task is commonly addressed from two perspectives: (a) unsupervised methods (e.g., \cite{roth2022towards,gudovskiy2022cflow}), which model normal data distributions to detect deviations, and (b) supervised methods (e.g., \cite{huang2024adapting,zhang2024mediclip,yao2023explicit}), which leverage labeled anomalies. For instance, PatchCore \cite{roth2022towards} compares test samples to a memory bank of normal embeddings, while CFLOW-AD \cite{gudovskiy2022cflow} models normality using normalizing flows and Gaussian distributions. Although unsupervised methods benefit from large datasets, real-world settings often involve few labeled anomalies, motivating the development of few-shot AD approaches. Yet, incorporating limited anomalies in training introduces reduced generalization. To mitigate this, DRA \cite{ding2022catching} learns disentangled representations, and BGAD \cite{yao2023explicit} employs boundary-guided semi-push-pull contrastive learning. 
                
                Recently, CLIP \cite{radford2021learning} has been leveraged for AD. April-GAN \cite{chen2023zero} maps CLIP-extracted visual features to the text feature space and uses multiple memory banks, which is an expensive setup with limited generalization. In medical AD, MVFA \cite{huang2024adapting} extends this by applying multi-level CLIP adaptation with fixed prompts. MediCLIP \cite{zhang2024mediclip}, instead, uses learnable prompts \cite{zhou2022coop} and synthetic abnormal data to improve generalization. However, a learnable prompt with OT/POT, as shown in our experiments, hinders performance, especially in certain modalities (e.g., histopathology). Our approach integrates POT and CL to enhance alignment and generalization across diverse medical data. \\

     \noindent \textbf{Optimal Transport.} Originally proposed for minimizing resource allocation costs \cite{villani2008optimal}, OT has become an important tool in machine learning for comparing probability distributions as feature sets \cite{peyre2019computational}.  
        OT has been used in tasks such as domain adaptation \cite{li2024unsupervised}, learning with noisy labels \cite{feng2023ot}, causal discovery \cite{tu2022optimal}, and federated learning \cite{li2024global}. 
        Partial OT extends classical OT by
        offering flexibility where exact mass preservation is impractical (See \ref{sec:method} for details). 
        PLOT \cite{chen2022plot} is the most relevant study, using OT to learn local prompts with CoOP \cite{zhou2022coop} for fine-grained vision-language matching. Motivated by PLOT, our approach uses POT instead and shows that POT alone is insufficient; CL is necessary for improved zero-shot and few-shot performance. This is the first study to apply OT and POT to medical AD while integrating CL, visual adapters, and learnable prompts for CLIP adaptation.

\section{Method}
\label{sec:method}
    In training, the data consists of triplets $\{x_i, c_i, M_i\}$, where each sample $x_i$ is an image of size $\mathbb{R}^{h \times w \times 3}$, the label $c_i$ takes a binary value, with 1 and 0 indicating normal and abnormal, respectively. When provided, the corresponding AS mask $M_i$ is a binary map of dimensions $h \times w$. At inference, given a test image $x_{\text{test}}$, the model outputs both an anomaly classification (AC) decision and an AS mask prediction.  
    Building on this setup, we propose \method (see Fig.~\ref{im:proposed}a), an innovative approach that leverages CLIP~\cite{radford2021learning}, incorporating visual features adaptation and projection alongside using POT for learning diverse text prompts. \method learns separate multi-modal representations for normal and abnormal samples by simultaneously integrating CL and POT. While CL ensures that prompts for each class are distinguishable, POT promotes diversity within learnable prompts for each class, leading to better distinctions between normal and abnormal classes (see Fig.~\ref{im:proposed}b). Below, we provide a detailed description of the components employed in the \method. 

   \noindent \textbf{Visual Feature Adapting.}
        To adapt the pre-trained CLIP vision encoder \cite{radford2021learning} for the AC and AS tasks in medical imaging, \method uses adapter and projection layers. This approach follows the findings of~\cite{huang2024adapting}, which shows that it is preferable to traditional fine-tuning, as it helps avoid overfitting due to high model complexity and limited data. We append a small set of learnable layers to the visual branches of CLIP while keeping the backbone frozen. 
        
        In detail, for an input image $x$, we extract the $i$-th layer \textit{feature} from the CLIP vision encoder, denoted as $f_{vis}^i(x) \in \mathbb{R}^{G \times d}$. Here, $G$ represents the grid size, $d$ is the feature dimension, and $i \in \{12,24\}$. Learnable adapter $A(.)$ and projector $P(.)$ consist of two transformation stages. The first stage addresses the domain gap between natural and medical images.
        By introducing a shared linear transformation layer \textit{$ W^i_{\text{shared}}$} that maps CLIP features into a feature space better suited for medical AD, formulated as  $F^i_{\text{shared}}(f_{vis}^i(x)) = \textit{\text{ReLU}}(W^i_{\text{shared}} f_{vis}^i(x))$.
        To effectively handle both AC and AS, the second stage focuses on extracting features specialized for each task by applying two distinct linear transformation heads on top of the shared features: the first head, dedicated to AC, captures high-level characteristics essential for detecting anomalies, while the second head, highlights fine-grained spatial details for AS, expressed as  
        $ F_{\text{Det}}^i(f_{vis}^i(x)) = \textit{\text{ReLU}}\Big(W^i_{\text{Det}}\, F_{\text{shared}}^i(f_{vis}^i(x))\Big) $  
        and  
        $ F_{\text{Seg}}^i(f_{vis}^i(x)) = \textit{\text{ReLU}}\Big(W^i_{\text{Seg}}\, F_{\text{shared}}^i(f_{vis}^i(x))\Big)$. As a result, two feature sets, $O_{Det}^i$ and $O_{Seg}^i $, are produced at each level.

        Both $A(.)$ and $P(.)$ follow the same transformation structure. However, $A(.)$ employs a residual connection as empirically found better-performing. Specifically, a constant value $\gamma$ is introduced as a residual ratio to balance knowledge preservation and adaptation, formulated as $ F_*^i(f_{vis}^i(x)) = \gamma F^i_{\text{shared}}(f_{vis}^i(x)) + (1-\gamma)f_{vis}^i(x) $, where $F_*^i$ serves as input to the $i+1$ layer of the vision encoder. \\
   
   \noindent \textbf{Learnable Prompt Sets and Fused Prompts.}
    Multiple learnable prompts (i.e., prompt sets) are used for each class (normal and abnormal) to support discriminative feature learning through POT. These prompt sets are fused to create class-discriminative embeddings via CL.
    The textual prompt for \textit{normal} class is defined as $ t_{\text{n}}^k = \{ v^k_1, v^k_2, \dots, v^k_L, cls_{\text{n}}^k \}$, where $cls_{\text{n}}^k$ represents the $k$'s normal class name (e.g., ``flawless'', ``healthy'')
    and $V_n^k = \{ v^k_l \}_{l=1}^{L}$ consists of learnable tokens, each with the same dimension as the original word embedding, 
        and $L$ denotes the number of context words. Given  $t_{\text{n}}^k$ as input, the CLIP text encoder ($f_{txt}$) generates a set of prompts for class normal, $ P_n = \{ f_{txt}(t_{\text{n}}^k) \}_{k=1}^{K}$,  where each prompt is in $\mathbb{R}^d$. 
        To enhance the learning of discriminative prompts across two classes of normal and abnormal, a fused prompt for the normal class is computed as    
        $ P_{\text{fused},n} = \frac{1}{K} \sum_{k=1}^{K} P_n^k $. The same procedure is applied to the abnormal class (with class names such as pathological, with anomalies), yielding $ P_{\text{ab}} = \{ f_{txt}(t_{\text{ab}}^k) \}_{k=1}^{K} $, and the corresponding fused prompt $ P_{\text{fused},ab} = \frac{1}{K} \sum_{k=1}^{K} P_{\text{ab}}^k $. \\
        
        \noindent \textbf{Multimodal Feature Alignment.}
        Let us first formally describe classical and partial optimal transport (OT and POT, respectively), and then explain how we perform multimodal feature alignment using POT.
        OT is a constrained optimization problem that aims to find the most efficient way to transfer probability mass between two distributions \cite{villani2008optimal}. In the discrete case, let $\alpha$ and $\beta$ be probability simplex vectors, and let $C \in \mathbb{R}^{|\alpha| \times |\beta|}$ be the cost matrix. The objective of OT is to determine the optimal transport plan $T$ by solving $dis_C (\alpha, \beta) = \min_{T \in U(\alpha, \beta)} \langle C, T \rangle$, where $\langle \cdot, \cdot \rangle$ denotes the Frobenius inner product, and $U(\alpha, \beta)$ represents the set of feasible transport plans: $ U(\alpha, \beta) = \left\{ T \in \mathbb{R}^{|\alpha| \times |\beta|}_+ \mid T \mathbf{1}_{|\beta|} = \alpha, \quad T^\top \mathbf{1}_{|\alpha|} = \beta \right\}. $ Directly solving the OT problem can be computationally expensive. To alleviate this, the Sinkhorn algorithm \cite{sinkhorn1967diagonal} introduces an entropic regularization term to accelerate the optimization process. The regularized OT formulation is given by: $ dis_C (\alpha, \beta) =\min_{T \in U(\alpha, \beta)} \langle C, T \rangle + \lambda \langle T, \log T \rangle$, where $\lambda \geq 0$ is a regularization parameter. Under this formulation, the optimal transport plan $T^*$ has a unique closed-form solution.
    
        POT extends the classical OT formulation by relaxing the strict equality constraints on marginal distributions \cite{frogner2015learning}. The objective function remains the same as OT formulation, but the feasible transport plan set $ U(\alpha, \beta) $ is modified to allow for mass variations: $ U(\alpha, \beta) = \left\{ T \in \mathbb{R}^{|\alpha| \times |\beta|}_+ \mid T \mathbf{1}_{|\beta|} \leq \alpha, \quad T^\top \mathbf{1}_{|\alpha|} = \beta \right\}.$ Here, satisfy $ \|\alpha\|_1 \geq \|\beta\|_1 = \textit{frac} $ with $ \textit{frac} \in [0,1] $. For computational efficiency, as in OT,  an entropic regularization term is added. The optimization problem can be reformulated as a Kullback-Leibler projection ~\cite{benamou2015iterative}, and as introduced in \cite{chang2023csot}, it can be solved in a few iterations by a fast implementation of Dykstra’s Algorithm \cite{dykstra1983algorithm} by only performing matrix-vector multiplications, which is very similar to the Sinkhorn Algorithm \cite{sinkhorn1967diagonal}. 
        
        In our implementation, by representing visual features and prompts as discrete distributions, POT maps prompts to relevant image patches. Unlike classical OT, which enforces strict alignment (mass conservation), POT relaxes this constraint. As a result, prompts are not forced to match every image patch, avoiding the inclusion of irrelevant information and improving focus on meaningful correspondences.
        
            In detail, to align visual feature sets $O_{Det}^i$ and $O_{Seg}^i$ with the prompt sets $P_n$ and $P_{ab}$ through POT, the cost matrix is defined based on the cosine distance between the visual feature sets and the prompt sets. For class $j \in {\{n, ab\}}$, the cost matrices at each layer are given by:  
            $
                C^i_{Det,j} = 1 - {O_{Det}^i}^\top P_j, \quad C^i_{Seg,j} = 1 - {O_{Seg}^i}^\top P_j
            $, where $C^i_{Det,j}$ and $C^i_{Seg,j} \in \mathbb{R}^{G \times k}$. 
            For multimodal alignment, a two-stage optimization strategy (i.e. inner and outer loop) is employed. In the inner loop, the optimal transport plans $T^{i,*}_{Det,j}$ and $T^{i,*}_{Seg,j}$ are driven based on the defined cost matrices. This is achieved while keeping all learnable parameters fixed. It results in transport distances $dis^i_{Det,j}$ and $dis^i_{Seg,j}$ $\in \mathbb{R}^{G \times k}$. In the outer loop, the transport plans are fixed, while the learnable parameters are optimized based on the distances as Eq. \ref{eq:score_det}(a), where $\tau$ stands for the temperature of the softmax. To compute the image-level score at each layer by POT $(\hat{c}^i_{pot})$, we aggregate the patch-level distances by summing over all $G$ number of patches and all $k$ number of prompts for each class.
            
            To align fused prompts $P_{fused,n}$ and $P_{fused,ab}$ with visual feature sets through CL, the cosine similarity ($sim$) of the prompts and visual features is calculated, resulting in patch-level scores. To obtain image-level scores, the scores of each patch are summed as Eq. \ref{eq:score_det}(b). The predicted AC score in the $i$-th  feature layer is $\hat{c}^i_j  = \hat{c}^i_{pot,j} +  \hat{c}^i_{cl,j}$.
           
            \newcommand{\meanG}{\mathop{\sum}\limits_{G \times k}}    \newcommand{\meanseg}{\mathop{\sum}\limits_{k}}
                        \newcommand{\meancl}{\mathop{\sum}\limits_{G}}
            \begin{equation}
            \label{eq:score_det}
            \resizebox{!}{1\baselineskip}{$
            \hat{c}^i_{pot,j}(y=j |x) = \frac{\exp\big(\meanG(1 - dis^i_{Det,j})/\tau\big)}{\sum_{j^\prime} \exp\big(\meanG(1 - dis^i_{Det,j^\prime})/\tau\big)}
            \text{ (a) , }
            \quad
            \hat{c}^i_{cl,j}(y=j |x) = \frac{1}{G}\meancl\big(\frac{\exp\big(sim(O^i_{Det},P_{fused,j})/\tau\big)}{\sum_{j^\prime} \exp\big( sim(O^i_{Det}, P_{fused, j^\prime})/\tau\big)}\big) \text{ (b) }
            $}
            \end{equation}
            The anomaly mask at each layer $(\hat{M}^i_{j})$ is generated by aggregating $(dis^i_{Seg,j} \in \mathbb{R}^{G \times k})$ across all $k$ prompts. The resulting logits from POT (i.e. $1 - dis^i_{Seg,j}$) and CL (i.e. $sim(O^i_{Seg}, P_{fused,j})$)  are subsequently aggregated. Aggregation at this stage has been empirically found to be more effective for AS (up to +30\% AD score), compared to the aggregation level similar to AC where aggregation is performed only at the final stage. Given that AS requires alignment with the input dimensions, \((\mathbf{BI})\) modifies the anomaly logits to a size of \(\sqrt{G} \times \sqrt{G}\) and subsequently resizes to the original input dimensions through bicubic interpolation (Eq. \ref{eq:score_seg}). \\
                 \begin{equation}
                \label{eq:score_seg}
                \resizebox{!}{1\baselineskip}{$
                \hat{M}^i_{j}(y=j |x) = \frac{\exp
                    \big(
                    \mathbf{BI}\big(\meanseg (1 - dis^i_{Seg,j}) +  sim(O^i_{Seg},P_{fused,j})\big) /\tau 
                    \big)}
                {\sum_{j^\prime} \exp
                    \big(
                    \mathbf{BI}\big(\meanseg (1 - dis^i_{Seg,j^\prime}) +  sim(O^i_{Seg},P_{fused,j^\prime})\big) /\tau 
                    \big)}
                $}
                \end{equation} 
                
        \noindent \textbf{Loss function.} The loss function at feature level $i$ is \( Loss^i = \omega_1\, \text{GDice}(\hat{M}^i, M) + \omega_2\, \text{Focal}(\hat{M}^i, M) + \omega_3\, \text{BCE}(\hat{c}^i,\, c) \).
        By default, all loss terms contribute equally ($\omega_1=\omega_2=\omega_3=1)$. $GDice(.,.)$, $Focal(.,.)$, and $BCE(.,.)$ correspond to Generalized Dice, Focal, and Binary Cross Entropy losses. GDICE loss is specifically designed for highly imbalanced segmentation tasks, and Focal loss is beneficial in class imbalance scenarios, which is satisfied at the pixel level AD, where anomalous pixels are significantly outnumbered by normal pixels. The overall loss is calculated as the sum of the losses at different feature levels: $Loss = \sum_{i} Loss^i$. 
        During inference, $\hat{c}^i$ and $\hat{M}^i$ are averaged across feature layers to compute the final output of the model.
        
\section{Experiments}

We used the BMAD benchmark \cite{bao2024bmad} in line with the latest SOTA: MVFA \cite{huang2024adapting}. It consists of five modalities and six datasets: brain MRI~\cite{baid2021rsna,bakas2017advancing,menze2014multimodal}, liver CT~\cite{bilic2023liver,landman2015miccai}, chest X-ray~\cite{wang2017chestx}, retinal OCT (composed of two datasets; OCT17~\cite{kermany2018identifying}, and RESC~\cite{hu2019automated}), and digital histopathology (HIS)~\cite{bejnordi2017diagnostic}. Brain, Liver, and RESC supply annotations for both AC and AS, while OCT17, Chest, and HIS are relevant only for AC. We report the results for AC and AS using the standard medical AD metric: Area Under the Receiver Operating Characteristic Curve (AUC). \\

\noindent \textbf{Implementation Details.}
We use the CLIP \cite{radford2021learning} with the ViT-L/14 backbone in line with the recent SOTA, such as~\cite{chen2023zero,huang2024adapting,zhang2024mediclip}. 
This CLIP model consists of 24 layers. We insert an adapter at the 12\textsuperscript{th} layer and attach the projector to the 24\textsuperscript{th} layer. This configuration is motivated by the ablation study in MVFA \cite{huang2024adapting}, which shows that using the 12\textsuperscript{th} layer yields better performance compared to alternatives such as the 6\textsuperscript{th}, 18\textsuperscript{th}, or their combinations.
The number of prompts in a prompt set is fixed to 4 as in PLOT \cite{chen2022plot}. Training is performed with the Adam optimizer with the learning rate of $1e^{-3}$, batch size 16, for 100 epochs. We set the adapter's residual ratio $\gamma$ to 0.2 by default. 
Regarding the hyperparameters for solving POT, we fix the entropic regularization weight $\lambda$ to 0.1, the maximum iteration number is set to 100, and the transport ratio \textit{frac} is set to different values: 0.4–0.9 (step 0.1), while the final model uses 0.8. An early stopping is employed using a threshold of $0.001$. \\
\begingroup
\renewcommand{\arraystretch}{1.2} % Adjust the row distance
\begin{table}[t]
\centering
\caption{ Comparisons with SOTA in terms of AUC (\%). Few-shot models use 16 samples per class. The results are reported in the AC/AS format. Best results are \textbf{bold}, second-best \underline{underlined}.}

\resizebox{1\linewidth}{!}{
\begin{tabular}{l|l|c|c|c|c|c|c|c}

 & Method  &  HIS  &  Chest  &  OCT17  &  Brain  &  Liver  &  RESC &  AVG   \\ \hline

\multirow{4}{*}{Unsup}
    &CFLOWAD \cite{gudovskiy2022cflow}  & 54.54 & 71.44 & 85.43 & 73.97/93.52 & 49.93/92.78 & 74.43/93.75 & 68.29/93.35 \\ 
    &RD4AD \cite{deng2022anomaly}  & 66.59 & 67.53 & 97.24 & 89.38/96.54 & 60.02/95.86 & 87.53/96.17 &  78.04/96.19\\ 
    &PatchCore \cite{roth2022towards}  & 69.34 & 75.17 & 98.56 & 91.55/96.97 & 60.40/96.58 & 91.50/96.39 & 81.09/96.65 \\ 
    &MKD \cite{salehi2021multiresolution} &  77.74 &  81.99 & 96.62 & 81.38/89.54 & 60.39/96.14 & 88.97/86.60 &  81.18/90.76\\ \hline

\multirow{6}{*}{Few-shot}

        & DRA \cite{ding2022catching}  & 79.16 & 85.01 & 99.87 & 82.99/80.45 & 80.89/93.00 & 94.88/84.01 & 87.13/85.82 \\
        
        & BGAD \cite{yao2023explicit}  & - & - & - & 88.05/95.29 & 78.79/99.25 & 91.29/97.07 & \hspace{0.3cm}-\hspace{0.2cm} /97.20   \\
        
        & APRIL-GAN \cite{chen2023zero}  & 81.16 & 78.62 &\underline{ 99.93 }& 94.03/96.17 & 82.94/{99.64} & 95.96/98.47 & 88.77/98.09 \\

        & MediCLIP \cite{zhang2024mediclip}  & 70.22 & 69.74 & 96.37 & 91.56/\textbf{98.08} & 79.31/98.95 & 86.51/94.07 & 82.28/97.03\\
        
        & MVFA \cite{huang2024adapting}  & \underline{82.62} & \underline{85.72} & 99.66 & \underline{94.40}/{97.70} & \underline{83.85}/\underline{99.73} & \underline{97.25}/\underline{99.07} & \underline{90.58}/\underline{98.83}\\
        
        & \method (Ours)  & \textbf{96.59} & \textbf{93.4} & \textbf{100} & \textbf{99.99}/\underline{97.97} & \textbf{97.62}/\textbf{99.84} & \textbf{99.4}/\textbf{99.3} &\textbf{ 97.83}/\textbf{99.03}  \\ \hline
\end{tabular}
}
%\vspace{-1em}
\label{tab:compare_all}
\end{table}
\begin{figure}[t!]
    \centering
    \includegraphics[width=0.8\linewidth]{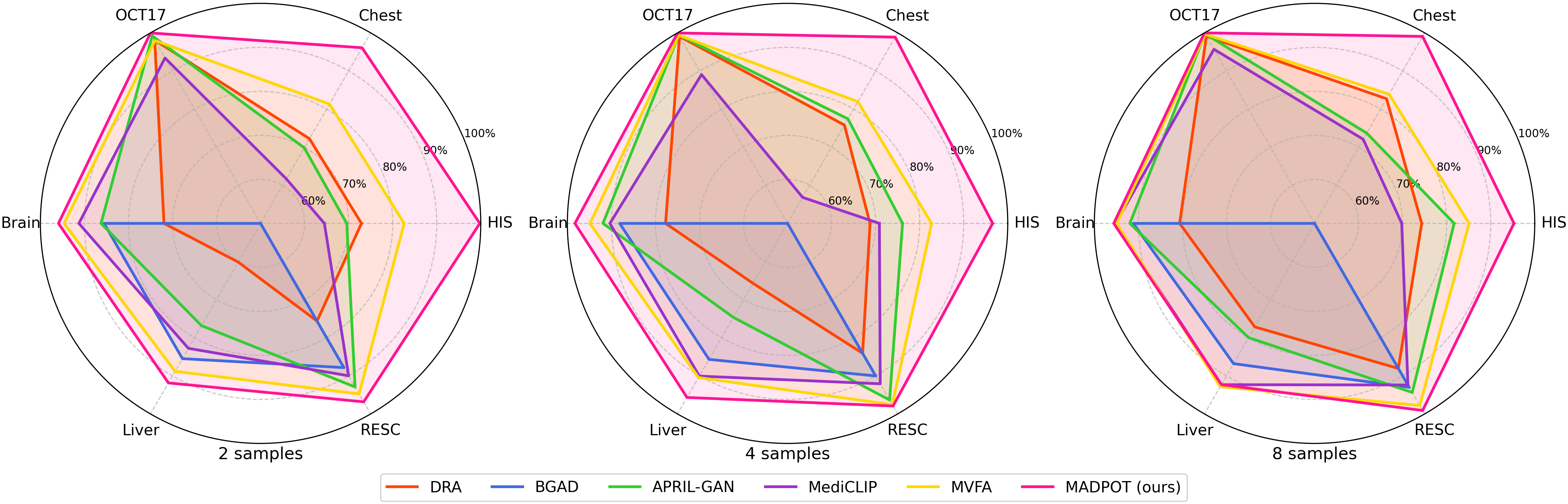}
 
    \caption{Comparison with few-shot SOTA for 2, 4, and 8 shots per class (AUC \%). For datasets with both AC and AS, the average of AC and AS is reported.}
    \label{im:various_shots}

\end{figure}

\noindent \textbf{Comparing Few-Shot Performance with SOTA.}
Table~\ref{tab:compare_all} presents a comparison of \method with SOTA methods. In that table, \textit{Unsup} methods rely on large normal training datasets, while few-shot methods use 16 normal and 16 abnormal examples. 
\method achieves the best AC performance across all six datasets, with an average AC score of 97.83\%, surpassing the second-best (i.e., MVFA~\cite{huang2024adapting}) by 7.25\%. Notably, on HIS and Chest, \method achieves significant gains of 13.97\% and 7.68\%, respectively, showing its ability on the AC task. For AS, \method achieves competitive performance, attaining near-perfect scores of 99.54\% on Liver and 99.3\% on RESC. On average, it outperforms all other methods, even though the second-best AS performance is close.

Moreover, Fig.~\ref{im:various_shots} compares few-shot methods with varying numbers of normal and abnormal samples. As visualized by the outermost pink area in all plots, \method consistently outperforms competing methods across all datasets, with particularly notable advantages in the Chest and HIS datasets. Although some methods, such as MVFA \cite{huang2024adapting}, show competitive performance in specific datasets, especially as the sample size increases to 8, none match \method across all. The consistent dominance of our approach confirms its value even when working with extremely limited training data.
\myvspace \\
\noindent \textbf{Cross-dataset analysis.} \method is evaluated against its strongest counterpart, MVFA~\cite{huang2024adapting}, to assess its generalization performance in a cross-dataset setting. Each model was trained on 16 samples from the datasets described above and evaluated on unseen target datasets of the same modality (see Table \ref{tab:xvalidation}). \method consistently outperforms MVFA in all datasets, achieving an average performance gain of 5.75\%, demonstrating its superior generalization. \\
\begin{table}[t!]
\centering
\caption{Cross-dataset evaluation. AC is reported for all datasets. Best are \textbf{bold}.}

\resizebox{\linewidth}{!}{
\begin{tabular}{lc c c c c c}
\toprule
{Source} & \multicolumn{2}{c}{Chest} & \xspace \xspace Brain \xspace \xspace  &  {\xspace \xspace OCT17 \xspace \xspace } & RESC \xspace \xspace  & \multirow{2}{*}{\xspace \xspace AVG} \\
\cmidrule(lr){2-3} \cmidrule(lr){4-4} \cmidrule(lr){5-6}
Target & NIHChest \cite{summers2019nih} & CheXpert \cite{irvin2019chexpert} & ADNI \cite{jack2008alzheimer} & \multicolumn{2}{c}{OCTDL \cite{kulyabin2024octdl}} &  \\ 
\midrule

MVFA \cite{huang2024adapting} 
& 61.91 & 80.41 & 53.94 & 88.47 & 93.94 & 75.73 \\

\method (Ours) & \textbf{64.9} & \textbf{90.69} & \textbf{54.63} & \textbf{99.21} &\textbf{ 97.95} & \textbf{81.48}  \\

\bottomrule
\end{tabular}
}
\label{tab:xvalidation}

\end{table}
\begingroup
\renewcommand{\arraystretch}{1.2} % Adjust the row distance
\begin{table}[t!]
\centering
\caption{ Comparisons with SOTA in terms of AUC (\%) in \textit{zero-shot} scenario. The results are reported in the AC/AS format. Best results are \textbf{bold}, second-best \underline{underlined}.}
\resizebox{\linewidth}{!}{
\begin{tabular}{l|l|c|c|c|c|c|c|c}

 & Method  &  HIS  & Chest  &  OCT17  &  Brain  &  Liver  & RESC  &  AVG   \\ \hline

\multirow{3}{*}{Zero-shot}                
        & APRIL-GAN \cite{chen2023zero} & 72.36 & 57.49 & 92.61 & 76.43/\underline{91.79} & 70.57/\underline{97.05} & 75.67/85.23 & 74.19/91.36 \\
        
        & MVFA \cite{huang2024adapting}  & \underline{77.9} & \underline{71.11} & \underline{95.4} & \underline{78.63}/90.27 & \underline{76.24}/\textbf{97.85} & \underline{83.31}/\textbf{92.05} & \underline{80.43}/\textbf{93.39}  \\
        
        & \method (Ours) & \textbf{90.35} & \textbf{76.81} &\textbf{ 95.9} & \textbf{94.24}/\textbf{93.81} & \textbf{88.21}/96.27 & \textbf{85.04}/\underline{86.6} &  \textbf{88.42}/\underline{91.56}\\ \hline
\end{tabular}
}

\label{tab:zero_results}
\end{table}

\noindent \textbf{Comparing Zero-Shot Performance with SOTA.}
Table \ref{tab:zero_results} presents a comparison of \method with zero-shot SOTA. 
\method consistently achieves superior performance in AC, surpassing the second best across all evaluated datasets with a notable average gain of 7.99\%. While our method's average AS score is marginally lower (by just 1.83\% than the top performer), this slight difference is only due to the RESC dataset, where \method lags by 6\% AUC.  \\

\noindent \textbf{Ablation studies.} 
\begingroup
\renewcommand{\arraystretch}{1.2} % Adjust the row distance
\begin{table}[t!]
\centering
\caption{Ablation study evaluating the impact of different prompt learning strategies in \textit{few-shot} scenario. Results are reported in the AC/AS format, with \textit{few-shot} referring to 16 samples per class. %\CB{we have to cancel zero shot no space, i will write it with one sentence instead, so don't do experiments..}
}
\label{tab:ablation_prompt_strategies}
\resizebox{\linewidth}{!}{
\begin{tabular}{c|ccc|c|c|c|c|c|c|c}
Scenario & \xspace CL \xspace &\xspace OT \xspace & \xspace POT \xspace & \xspace HIS \xspace & \xspace Chest \xspace & \xspace OCT17 \xspace & \xspace Brain \xspace & \xspace Liver \xspace & \xspace RESC \xspace & \xspace AVG \xspace  \\ \hline
\multirow{5}{*}{few-shot} 
&\ding{55}&\ding{55}&\ding{55}& \underline{ 86.87} & \underline{88.26} & 99.71 &93.91/95.73 & 92.35/{99.62}& 96.33/98.46 & 92.90/97.94\\
&\ding{51}&\ding{55}&\ding{55}& 85.61 & 89.11 & 99.47 &94.13/\underline{97.24} & 92.35/\underline{99.78} & 97.45/\underline{99.07} & 93.02/\underline{98.70}  \\
&\ding{55}&\ding{51}&\ding{55} & 9.8 & 46.55 & 75.43 & 15.75/51.9 & 55.01/53.05 & 52.59/52.05 &	42.52/52.33 \\
&\ding{55}&\ding{55}&\ding{51} & 67.54 & 48.25 & 83.34 & 20.36/49.76 & 61.09/49.87 & 52.49/50.42 & 55.51/50.02 \\
&\ding{51}&\ding{51}&\ding{55} & 79.05 & 88.25 & \textbf{100} &\underline{ 99.95}/95.90 & \underline{94.07}/99.15 &\textbf{ 99.85}/98.87 & \underline{93.53}/97.97 \\
&\ding{51}&\ding{55}&\ding{51} & \textbf{96.59} & \textbf{93.4} & \textbf{100} &\textbf{ 99.99}/\textbf{97.97 }& \textbf{97.62}/\textbf{99.84}	& \underline{99.40}/\textbf{99.30}	& \textbf{97.83}/\textbf{99.03}\\ \hline
\end{tabular}
}

\end{table}
\begin{comment} 
\hline
\multirow{5}{*}{zero-shot} 
&\ding{55}&\ding{55}&\ding{55}& 78.81 & 49.25 & - &	70.45/\textbf{88.17} & \underline{72.12}/\textbf{95.81 }& 71.58/\underline{79.34 }& 68.44/\textbf{87.77}\\
&\ding{51}&\ding{55}&\ding{55}&   76.68 & 40.52 & - & 75.42/86.59 & 76.34/91.36 &70.69/\textbf{83.5} & 67.93/\underline{87.15}\\
&\ding{55}&\ding{51}&\ding{55} & 80.13 & 59.37 & - & 60.51/50.9 & 58.52/53.48 &  \underline{85.15}/51.03 &	68.74/51.80 \\
&\ding{55}&\ding{55}&\ding{51} & 91.55 & \textbf{59.35}&	- & \underline{92.35}/49.44 & 47.65/52.06 & 64.48/50.13 & 71.08/50.54  \\
&\ding{51}&\ding{51}&\ding{55} & \textbf{98.31} & 46.56 & - & 80.58/78.86 & 70.48/93.94 & \textbf{85.52}/76.49 & \underline{76.29}/83.1 \\
&\ding{51}&\ding{55}&\ding{51} & \underline{96.84} & \underline{49.81} & - & \textbf{95.27}/\underline{87.28} & \textbf{72.87}/\underline{93.96 }& 82.81/76.9 & \textbf{79.52}/86.05\\ \hline
\end{comment}
\begingroup
\renewcommand{\arraystretch}{1.2} % Adjust the row distance
\begin{table}[t!]
\centering
\caption{Ablation study evaluating the impact of adapter and projector. Results are reported in the AC/AS format, with \textit{few-shot} referring to 16 samples per class.}
\label{tab:ablation_layers}
\resizebox{\linewidth}{!}{
\begin{tabular}{c|cc|c|c|c|c|c|c|c}
Scenario & \xspace Adapter \xspace &\xspace Projector \xspace & \xspace HIS \xspace & \xspace Chest \xspace & \xspace OCT17 \xspace & \xspace Brain \xspace & \xspace Liver \xspace & \xspace RESC \xspace & \xspace AVG \xspace   \\ \hline
\multirow{5}{*}{few-shot} 

&\ding{55}&\ding{55}&  79.72 & 85.74 & \underline{97.69} & 88.57/96.33 & 67.25/98.05 & 97.12/\underline{96.44} & 86.01/\underline{96.94} \\
&\ding{51}&\ding{55} & \textbf{97.71 }& \underline{86.26} & 84.85&\textbf{ 100}/\underline{96.62} &\underline{96.57}/\underline{98.95} & 51.58/89.25 & \underline{86.16}/94.94  \\
&\ding{55}&\ding{51} & 58.87 & 52.52 & 70.09 & 99.59/91.85 & 86.8/96.24 & \underline{97.5}/\underline{96.44} &	77.56/94.84 \\
&\ding{51}&\ding{51} & \underline{96.59} & \textbf{93.4} & \textbf{100} & \underline{99.99}/\textbf{97.97} & \textbf{97.62}/\textbf{99.84 }& \textbf{99.40}/\textbf{99.30} &\textbf{ 97.83}/\textbf{99.03}  \\ \hline
\end{tabular}
}

\end{table}
Several ablation studies were conducted to evaluate different components of our approach. This includes assessing the impact of prompt learning strategies in the text branch (Table~\ref{tab:ablation_prompt_strategies}), evaluating the effectiveness of projection and adapter layers in the vision branch (Table~\ref{tab:ablation_layers}), and the effect of POT parameter \textit{frac}. Table~\ref{tab:ablation_prompt_strategies} presents six variations based on the \textbf{inclusion or exclusion of CL, OT, POT, and their combinations} (e.g., CL with OT or CL with POT). Note that as POT is derived from OT, they cannot be used in combination. The case where all components are excluded (first row in the table) corresponds to using fixed, hand-crafted prompts, specifically adopting those from MVFA~\cite{huang2024adapting}. The case where only CL is enabled corresponds to using learnable prompts following the CoOP approach~\cite{zhou2022coop}.

Our analysis reveals several key findings. \textit{i)} Learnable prompts outperform fixed prompts, and the combination of CL and POT achieves the highest performance. \textit{i)} OT or POT alone leads to significant performance degradation, demonstrating their inability to learn effective prompts without CL. \textit{iii)} While CL alone improves AS (98.7\% vs. 97.94\%), the addition of POT provides the most balanced improvement, boosting AC by 4.93\% compared to fixed prompts without compromising AS performance. \textit{iv)} OT/POT shows clear benefits in AC, with gains also evident in AS. 

Table~\ref{tab:ablation_layers} evaluates \textbf{the impact of the adapter and projector}, showing that using the CLIP vision encoder without any adaptation (i.e., the first row) leads to suboptimal performance. This highlights the necessity of adaptation to better align the vision encoder with the AD task. Among the adaptation methods, the adapter was found to outperform the projector, while their combination achieved the best performance, with gains of 11.67\% in AC and 4.19\% in AS over the adapter-only setting. When considering the results in Table~\ref{tab:ablation_prompt_strategies} and Table~\ref{tab:ablation_layers} together, it becomes clear that removing vision adaptation while retaining prompt learning with POT and CL yields lower results, particularly for AC, compared to the opposite configuration. 

\textbf{The impact of \textit{frac}} on AC and AS was studied across three datasets: Brain, Liver, and RESC. Results showed sensitivity to this hyperparameter, with optimal performance between 0.5 and 0.8 (AC: 91.94\%-99.00\%, AS: 96.79\%-99.03\%). Extreme values (\textit{frac} = 0.4 or 0.9) caused significant degradation (AC/AS: 81.61\%/95.21\% for 0.4, 82.76\%/95.53\% for 0.9). Peak performance was achieved at 0.8, suggesting that excluding 20\% of the mass enhances relevant features.

\section{Conclusion}
We presented \method, which adapts CLIP to medical AD by integrating learnable prompt tuning with POT and CL, and incorporating adaptive vision components for both semantic understanding and pixel-level segmentation in few/zero-shot settings. \method outperforms baselines across diverse datasets, demonstrating robustness and generalization to varying modalities and distributions. 
Ablation studies show that combining learnable prompts with CL and POT is most effective in the text branch, while the adapter–projector pair drives vision performance, with notable gains. Their joint use further boosts results. Future work will extend \method to support temporal data processing.

\begin{credits}
\subsubsection{\ackname} This preprint has not undergone peer review or any post-submission revisions or corrections. The Version of Record of this contribution has been accepted to the 23rd International Conference on
Image Analysis and Processing (ICIAP) 2025; the DOI can be found in the arXiv comment section.
We acknowledge the financial support of the PNRR project FAIR - Future AI Research (PE00000013), under the NRRP MUR program funded by the NextGenerationEU.

%\subsubsection{\discintname}
%The authors have no competing interests to declare that are relevant to the content of this article.
\end{credits}

\bibliographystyle{splncs04}
% \bibliography{main}
\bibliography{bib2}

\end{document}